\documentclass[10pt,twocolumn,letterpaper]{article}

\usepackage{cvpr}              
\usepackage{colortbl}  
\usepackage{array}   
\usepackage{booktabs}
\usepackage{multirow}
\usepackage{amssymb}
\usepackage{bbding}
\usepackage{pifont}
\usepackage{array}
\newcommand{\myfontsize}{\fontsize{8pt}{10pt}\selectfont}
\definecolor{LightGray}{gray}{0.9}
\usepackage[dvipsnames]{xcolor}
\newcommand{\red}[1]{{\color{red}#1}}

\definecolor{cvprblue}{rgb}{0.21,0.49,0.74}
\usepackage[pagebackref,breaklinks,colorlinks,citecolor=cvprblue]{hyperref}

\def\paperID{909} 
\def\confName{CVPR}
\def\confYear{2024}

\title{ODCR: Orthogonal Decoupling Contrastive Regularization \\
for Unpaired Image Dehazing}

\author{Zhongze Wang\quad Haitao Zhao\thanks{Corresponding author.}\quad Jingchao Peng\quad Lujian Yao \quad Kaijie Zhao\\
East China University of Science and Technology, Shanghai, China\\
{\tt\small \{zzwang, lujianyao, kjzhao\}@mail.ecust.edu.cn, haitaozhao@ecust.edu.cn, starry-sky@outlook.com}}

\begin{document}
\maketitle
\vspace*{-10pt}
\begin{abstract}
    Unpaired image dehazing (UID) holds significant research importance due to the challenges in acquiring haze/clear image pairs with identical backgrounds. This paper proposes a novel method for UID named Orthogonal Decoupling Contrastive Regularization (ODCR). Our method is grounded in the assumption that an image consists of both haze-related features, which influence the degree of haze, and haze-unrelated features, such as texture and semantic information. ODCR aims to ensure that the haze-related features of the dehazing result closely resemble those of the clear image, while the haze-unrelated features align with the input hazy image. To accomplish the motivation, Orthogonal MLPs optimized geometrically on the Stiefel manifold are proposed, which can project image features into an orthogonal space, thereby reducing the relevance between different features. Furthermore, a task-driven Depth-wise Feature Classifier (DWFC) is proposed, which assigns weights to the orthogonal features based on the contribution of each channel's feature in predicting whether the feature source is hazy or clear in a self-supervised fashion. Finally, a Weighted PatchNCE (WPNCE) loss is introduced to achieve the pulling of haze-related features in the output image toward those of clear images, while bringing haze-unrelated features close to those of the hazy input. Extensive experiments demonstrate the superior performance of our ODCR method on UID.
\end{abstract}
    
\vspace{-10pt}
\section{Introduction}
\label{sec:intro}

\begin{figure*}
  \centering
  \includegraphics[width=\linewidth]{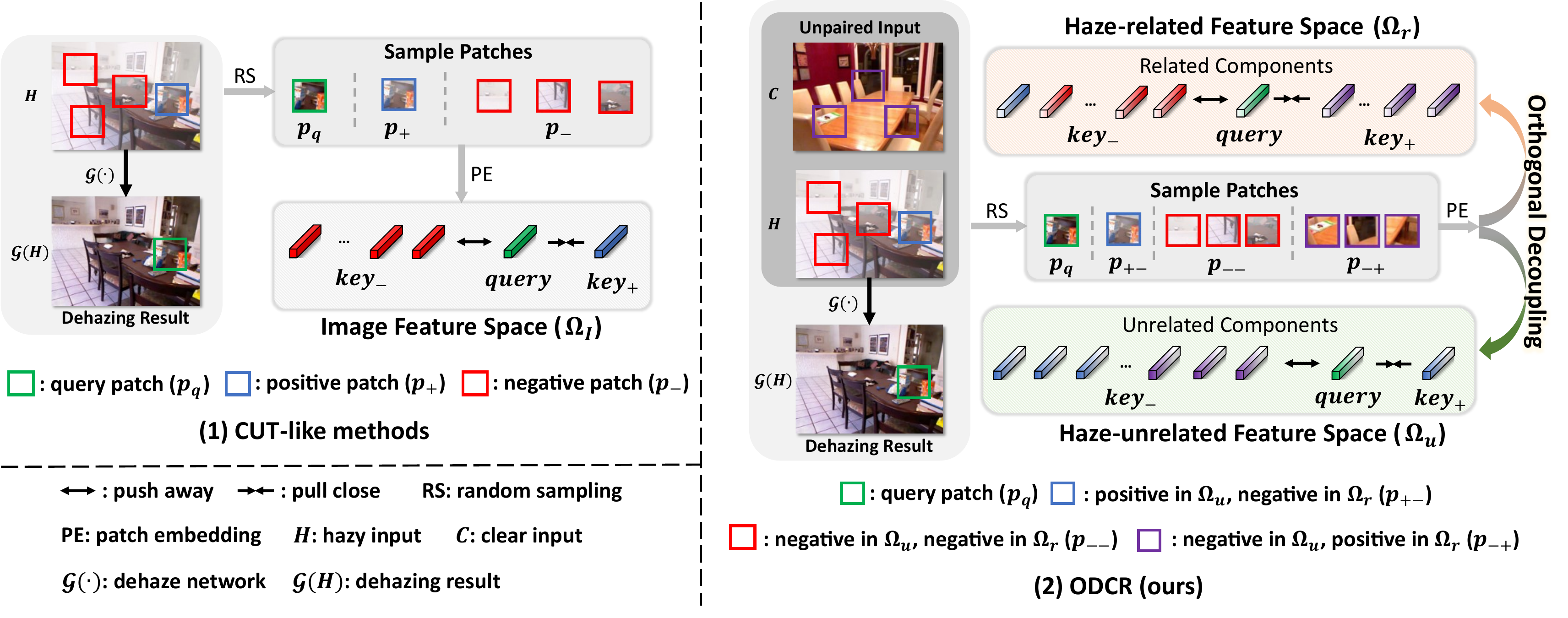}
  \caption{Illustration of how (a) the CUT-like methods and (b) ODCR work. CUT-like methods directly pull features of the query patch and positive patch close, leading to a contradiction in maximizing the mutual information between the two patches and dehazing. In ODCR, orthogonal decoupling are proposed to decouple image features to haze-related (describing haze level) and unrelated (describing non-haze information, such as semantic and texture) components. Then the mutual information between query and the positive patches in different feature spaces are maximized, thus avoiding the above contradiction.}
  \label{fig:intro}
  \vspace*{-10pt}
  \end{figure*}

Haze is a common atmospheric phenomenon caused by the accumulation of aerosol particles. It can cause severe quality degradation of images, which can affect subsequent computer vision tasks \cite{zhang2020unified, li2023detection}. The degradation of haze effect can be described as the atmospheric scattering model (ASM) \cite{drake1985mie,mccartney1976optics}:

\begin{equation}\label{eq:1}
    \textbf{I}(x)=\textbf{J}(x)t(x)+\textbf{A}(1-t(x))
\end{equation}
where $\textbf{I}(x)$, $\textbf{J}(x)$, $t(x)$ and $\textbf{A}$ stand for the hazy image, the clear image, the transmission map (T-map) and the global atmospheric light separately. 

To enhance image clarity and detail, numerous image dehazing methods have been introduced. Early image dehazing methods \cite{he2010single,zhu2015fast,fattal2014dehazing,berman2016non,berman2018single} are mainly based on handcrafted priors. These methods conduct statistical analyses on hazy and clear images to acquire prior knowledge for image enhancement, but have limited robustness due to specific assumptions. With the development of deep neural networks, many deep learning dehazing methods \cite{cai2016dehazenet,li2017aod,deng2020hardgan,yeperceiving,wu2021contrastive,song2023vision} have emerged, which train networks in a supervised manner on large-scale synthetic datasets, resulting in significantly improved dehazing performance.

Despite their impressive performance on synthetic data, the real-world applicability of these methods is limited by the challenge of acquiring paired images with identical backgrounds. Consequently, research into training dehazing models on unpaired datasets is gaining traction. Most current unpaired image dehazing (UID) strategies \cite{yang2022self,engin2018cycle,chen2022unpaired} adopt the Cycle-GAN framework \cite{zhu2017unpaired}, which constructs hazy-clear-hazy and clear-hazy-clear conversion cycles. These approaches depend on cycle-consistency loss to ensure content consistency across the dehazing process. However, the cycle-consistency loss assumes a bijective relationship between the two domains \cite{park2020contrastive}, which is too strict for image dehazing. In real world, a clear image can correspond to a hazy image with varying degrees of haze within the same scene.

To bypass this bijection limitation, CUT-like methods \cite{park2020contrastive, luo4620491farewell,wang2024ucl} have been introduced, eschewing the Cycle-GAN architecture for a singular GAN framework. These methods preserve consistency by maximizing mutual information between the features of a query patch in the dehazed output and the corresponding patch in the original hazy input, as depicted in Fig. \ref{fig:intro} (a). Nonetheless, this approach incurs a contradiction between maximizing mutual information and attaining effective dehazing, and do not fully utilize the guiding role of clear images for dehazing.

To achieving background consistency and bridging the gap between hazy and clear domains, one feasible idea involves the decoupling of query patch features into haze-related components, which quantify the level of haze, and haze-unrelated components, embodying non-haze attributes like semantics and texture. The objective is to maximize the mutual information between corresponding haze-related components of the query patch and clear image patches, as well as between haze-unrelated components of the query patch and the hazy patch at the same location.

The above feature decoupling faces two challenges. The first is \textbf{how to decouple the image features to components with low relevance.} The inherent blending of haze-related and unrelated features underlies the conflict between maximizing mutual information and effective dehazing. Decoupling into components of reduced relevance is crucial to resolving this conflict. According to the ASM \cite{drake1985mie,mccartney1976optics}, hazy images are captured by deep coupling of multiple physical quantities, making it difficult to completely realize the decoupling of features. And the second challenge is \textbf{how to assign the decoupled features as haze-related/-unrelated components.} Without the guidance of ground truth images, it becomes difficult for networks to distinguish between features pertaining to haze and those that do not.

To address above challenges, a novel Orthogonal Decoupling Contrastive Regularization (ODCR) is proposed for UID in this paper. In ODCR, we first repartition the samples according to different feature spaces as shown in Fig. \ref{fig:intro} (b). To solve the first challenge, we propose to introduce orthogonal constraint, which is widely used in traditional machine learning \cite{wold1987principal,bro2014principal,zhao2006novel} and deep learning \cite{cui2021multitask,liu2023learning,hu2021dynamic,peng2021second}, to decouple of image features into components with low relevance. To address the second challenge, a self-supervised Depth-wise Feature Classifier (DWFC) for mapping image features to hazy or clear labels is introduced. DWFC yields heat vectors that reflect the significance of each channel in discerning whether the feature is extracted from a hazy or clear image can be obtained. Based on the sample repartition and heat vectors, a Weighted PatchNCE (WPNCE) loss is proposed to realize the pulling of related/unrelated features in different feature space.

In summary, our main contributions are as follows:

\begin{itemize}
  \item The proposed ODCR projects image features to orthogonal space by Orthogonal-MLPs which are geometrically optimized on the Stiefel manifold to reduce the relevance between features.
  \item A self supervised DWFC is proposed to assign orthogonal features to haze-related and unrelated components, which provides weights indicating the degree of relevance of each channel to haze is proposed. 
  \item A Weighted PatchNCE is proposed to maximize the mutual information between the corresponding components of query and positive samples in different feature spaces.
\end{itemize}

\section{Related Works}
\label{sec:RelatedWorks}

\textbf{Unpaired Image Dehazing.} 
Considering the difficulty of obtaining large-scale data for supervised training, some methods \cite{zhao2019dd,yang2022self,chen2022unpaired1,engin2018cycle,chen2022unpaired} focus on learning mappings that restore hazy images to clear images from unpaired data. Zhu \etal \cite{zhu2017unpaired} first propose Cycle-GAN network based on cycle-consistency loss to solve the unpaired image-to-image (i2i) problem. Engin \etal \cite{engin2018cycle} proposed a Cycle-GAN-like method, which combines cycle-consistency and perceptual loss for UID. Afterwards, some Cycle-GAN-like UID methods \cite{chen2022unpaired1,zhao2019dd,chen2022unpaired,yang2022self} are proposed, among which $D^4$ proposed by Yang \etal \cite{yang2022self} realizes density-awareness by decomposing the transmission map in ASM into haze density and background depth, and then achieves excellent results on multiple datasets. Chen \etal \cite{chen2022unpaired} propose CDD-Net combining adversarial contrastive loss and cycle-consistency loss for extracting task-relevant and -irrelevant information. However, these Cycle-GAN-like methods are all based on the bijection assumption between the haze and clear domains \cite{park2020contrastive}, which is overly strict as a scene may correspond to multiple levels of haze.

To avoid the problems associated with the bijection assumption, Park \etal \cite{park2020contrastive} propose CUT, which maintains consistency by maximizing the mutual information between patches at the same location in the input and output. However, in image dehazing, the contradiction arises between maximizing the mutual information between the hazy-clear patch pair and the requirement that the output be a haze-free image. Subsequent CUT-like methods \cite{luo4620491farewell,wang2024ucl} do not focus on this contradiction either. Unlike these methods, our ODCR mitigates the contradiction by decoupling features into haze-related and unrelated components, individually aligning them with the corresponding components of the clear image and the original input, respectively.
\noindent
\textbf{Orthogonal Constraint.}
Orthogonal constraints play a pivotal role in diminishing feature relevance and curtailing redundant information in traditional machine learning \cite{wold1987principal, bro2014principal, zhao2006novel} and deep learning \cite{cui2021multitask, liu2023learning, hu2021dynamic, peng2021second}. While reduced rank Procrustes rotation \cite{zhao2020sparse} and eigen decomposition \cite{li2022robust} address orthogonal constraint problems in traditional machine learning, these methods are not applicable within deep learning frameworks. For deep learning methods with orthogonal constraint, one way is to convert the problem to an unconstrained one using Lagrange multipliers. However, the above method views the problem as a "black box" and it is hard to take advantage of orthogonal spaces \cite{nocedal1999numerical}. Some methods \cite{cui2021multitask,liu2023learning} include orthogonal regularizations in loss function, which do not guarantee that the parameters are in orthogonal space. In contrast to the aforementioned methods, our ODCR proposes to solve the orthogonal constraint problem using geometric optimization on the Stiefel manifold and thus perform a strict orthogonal decoupling of image features.

\noindent
\textbf{Contrastive Learning for Image Dehazing.}
Contrastive learning \cite{chen2020simple,He_2020_CVPR,han2020self,hjelm2018learning,caron2018deep,he2019rethinking,zbontar2021barlow} has shown power in high-level self-supervised representation learning tasks. In recent years, contrastive learning are also applied to low-level image enhancement tasks \cite{wu2021contrastive,wang2023uscformer,zheng2023curricular,wang2023restoring,zou2022estimating,chen2022unpaired2}. Wu \etal \cite{wu2021contrastive} first used contrastive learning in image dehazing and the proposed AECR aligns the features of the generated query image with those of the ground truths at pixel level. Zheng \etal \cite{zheng2023curricular} propose to additionally bound the solution space with results of other methods, considering that the lower bound of the solution space is always far away from positive samples. The above methods are applications of contrastive learning on supervised image dehazing. Chen \etal \cite{chen2022unpaired} introduce adversarial contrastive learning in CycleGAN network to disentangle task-relevant and -irrelevant factors for UID. However, the combination of GAN and adversarial contrastive learning makes the training process unstable. CUT-like methods \cite{park2020contrastive,luo4620491farewell,wang2024ucl} are based on PatchNCE, which is a contrastive regularization that pulls the query patch in the output close to the patch at the same location in the hazy input, leading background consistency. But it produce a contradiction of pulling the query to hazy or clear. In ODCR, we propose a Weighted PatchNCE (WPNCE), which avoids the contradiction by maximizing the mutual information of haze-related/-unrelated components of query and key sample features, respectively.
\section{Orthogonal Decoupling Contrastive Regularization}

In training process, given two unpaired clear image set $\mathcal{X}_C=\{C_i\}_{i=1}^{N_C}$ and hazy image set $\mathcal{X}_H=\{H_j\}_{j=1}^{N_H}$, a couple of unpaired images $\{C, H\}$ is input and we aim to train a generator $\mathcal{G} $ that can output a clear image $\mathcal{G}(H)$, which has a haze degree that converges to a clear image and keeps the haze-unrelated information such as image texture and semantics consistent with $H$.

\subsection{Sample Repartition} \label{Sec:3.2}
In the CUT-like methods depicted in Fig. \ref{fig:intro}, a query patch $p_q$ from the generated dehazed image is paired with a corresponding patch $p_+$ at the same location in the input hazy image as a positive sample, while other patches in the input image serve as negative samples $p_-$. This approach to sample partitioning, however, exhibits two critical limitations. First, it overlooks the influence of the clear domain, which is essential for restoring the haze level in the output to that of a clear image. Second, it creates an inherent conflict in determining whether to align the haze level of $p_q$ close to that of $p_+$, when attempting to increase the mutual information between the positive sample pairs.

To overcome the identified limitations, we propose a refined strategy for partitioning the positivity and negativity of sample patches. We assume that the features of a patch contain both haze-related features describing the haze level and haze-unrelated features containing image texture, semantics. For any given patch, its haze-related and unrelated components are separately classified as positive or negative within their respective feature spaces, as illustrated in Fig. \ref{fig:intro} (b). We adopt a dual subscript system to categorize the nature of samples. The first subscript indicates positivity or negativity in terms of haze-unrelated features: a patch at the same location as the query patch in the hazy domain $H$ is deemed positive, while all others are negative. The second subscript signifies positivity or negativity concerning hazy or clear: patches in $H$ are negative, whereas those in the clear domain $C$ are positive. This methodology results in distinct notational representations for various patches, as elaborated below:
\begin{itemize}
    \item $p_{_{+-}}$: the patch in $H$ with the same position as $p_q$;
    \item $p_{_{-+}}$: all patches in $C$;
    \item $p_{_{--}}$: all patches other than $p_{_{+-}}$ in $H$.
\end{itemize}

\begin{figure}
\centering
\includegraphics[width=\linewidth]{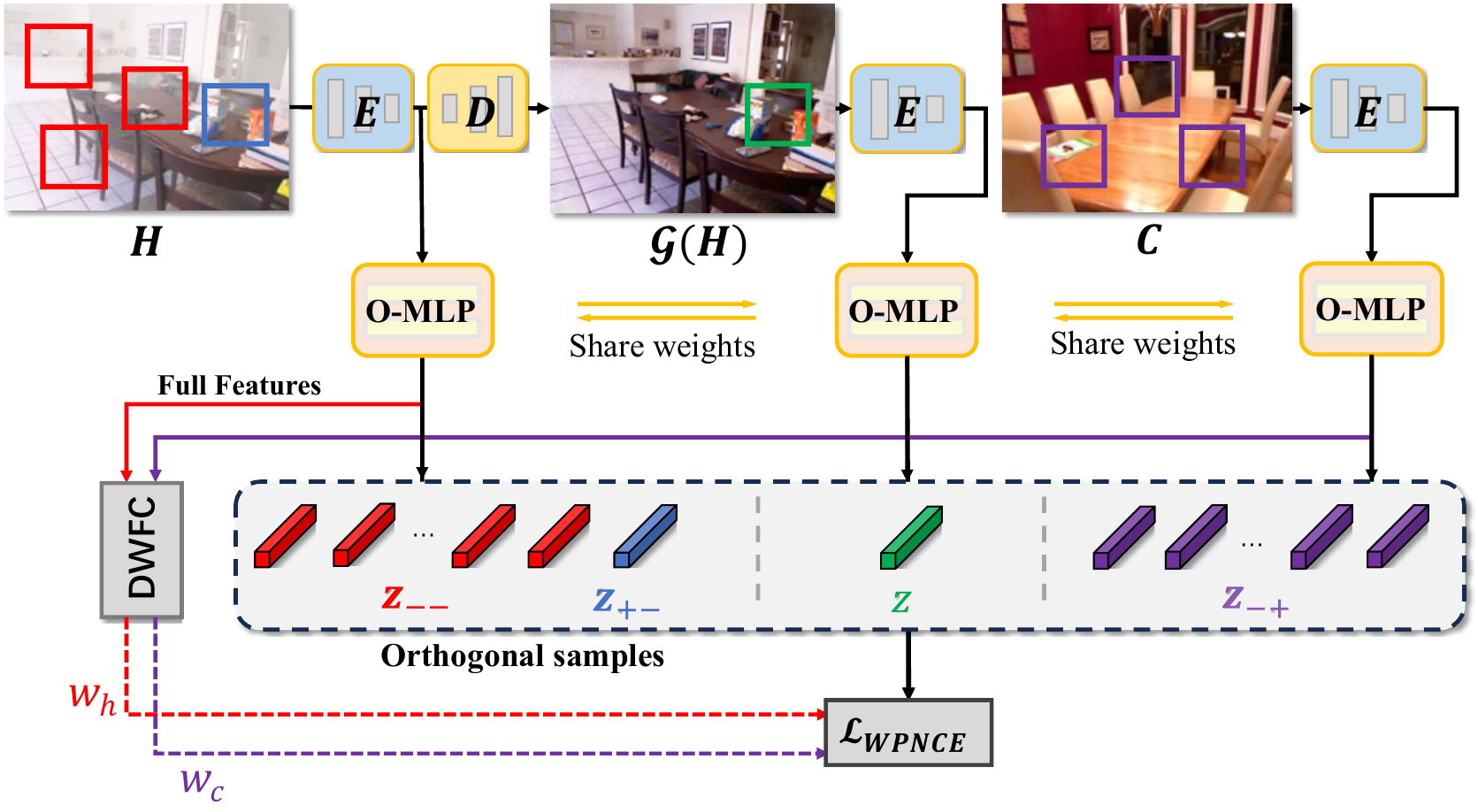}
\caption{The pipeline of the proposed ODCR.}\label{fig:Pipeline}
\vspace*{-10pt}
\end{figure}


\subsection{Orthogonal Decoupling}
In this subsection, we provide an introduction to how ODCR achieves orthogonal decoupling and solve the two chanllenges mentioned in Sec. \ref{sec:intro}. 
\vspace*{-10pt}
\subsubsection{Orthogonal Projection}
\textbf{Orthogonal MLP.} To achieve decoupling of haze-related and unrelated features, the relevance between the two kind of features needs to be reduced. Therefore, a MLP with orthogonal constraints is proposed to project the image features into the orthogonal space to reduce the relevance between the features:
\begin{equation}\label{eq:2}
    z_{_k}=\mathcal{H} _\Theta(\mathcal{G}_{enc}^i(p_{_k})), s.t. \Theta^T\Theta=I
\end{equation}
where $\mathcal{H} _\Theta(\cdot )$ stands for the MLP with orthogonal constraint and $\Theta$ stands for its parameter matrix. $\mathcal{G}_{enc}^i(\cdot )$ represents the feature of the $i$-th encoder layer in the generator $\mathcal{G}$. 

To solve the problem with orthogonal constraints, one way is to convert it to an unconstrained problem using Lagrange multipliers \cite{bertsekas2014constrained}. However, the method views the problem as a "black box" and it is hard to take advantage of orthogonal spaces \cite{nocedal1999numerical}. Some methods \cite{cui2021multitask,liu2023learning} include orthogonal regularizations in loss functions, which cannot guarantee that the parameters are in orthogonal space. Therefore, we propose to solve the orthogonal constraint problem using geometric optimization on the Stiefel manifold and thus perform a strict orthogonal decomposition of the features.

\begin{figure}[t]
    \centering
    \includegraphics[width=\linewidth]{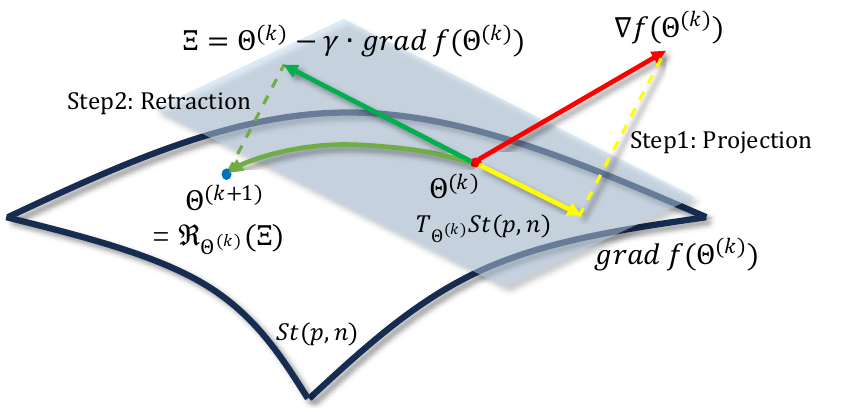}
    \caption{The illustration of the geometric optimization on the Stiefel manifold.}\label{fig:GeometricOptimization} 
    \vspace*{-10pt}
\end{figure}

\noindent
\textbf{Geometric Optimization on the Stiefel Manifold.} 
A Stiefel manifold is a set containing all orthogonal matrices in a specified space, i.e.
$St(p,n)\triangleq \{\Theta \in \mathbb{R} ^{n\times p}: \Theta ^T\Theta =I_p\}$. Its tangent space at a point $\Theta$ can be defined as: $T_\Theta St(p,n)\triangleq \{Z \in \mathbb{R} ^{n\times p}: \Theta^TZ+Z^T\Theta=0\}$. For our orthogonal decomposition problem, the ideal approach is to find the optimal solution of $\mathcal{H} _\Theta$ on the Stiefel manifold. 

Assuming that $f(\Theta)$ is a loss function defined in the Euclidean space and $\nabla f(\Theta)$ is its gradient in the Euclidean space, it cannot be optimized directly using optimizers such as SGD \cite{ruder2016overview} and ADAM \cite{kingma2014adam}, but requires an additional two process. Denote the Riemannian gradient $grad\ f(\Theta)$ as the tangent vector gradient of $f(\cdot )$ on the tangent space at point $\Theta$:
\begin{equation} \label{eq:4}
    grad\ f(\Theta)=\nabla f(\Theta)-\frac{1}{2}\Theta\Theta^T\nabla f(\Theta)-\frac{1}{2}\Theta{\nabla f(\Theta)}^T\Theta
\end{equation}
which points to the direction where the loss function $f(\cdot)$ on the Stiefel manifold ascends steepest and it can be proved by the following theorem:

\noindent
\textbf{Theorem 1.} Given $\nabla f(\Theta)$ in Euclidean space and $grad\ f(\Theta)$ defined by Eq. \ref{eq:4}, $grad\ f(\Theta)$ is the orthogonal projection of $\nabla f(\Theta)$ onto the tangent space of the Stiefel manifold.
And the first step is:

\noindent
\textbf{Step 1:} Project the gradient of the Euclidean space onto the tangent space the Stiefel manifold. And the updating process on the tangent space $T_{\Theta^{(k)}} St(p,n)$ at iteration k+1 is:
\begin{equation}
    \Xi =\Theta^{(k)}-\gamma grad\ f(\Theta^{(k)})
\end{equation}
where $\gamma$ is the gradient update step size. After that, the point $\Xi$ on the tangent space need to be remapped onto the Stiefel manifold based on the following theorem:

\noindent
\textbf{Theorem 2.} Given a point $\Xi=\Theta^{(k)}-\gamma grad\ f(\Theta^{(k)})$ on the tangent space $T_{\Theta^{(k)}} St(p,n)$ of a point on a Stiefel manifold, there is a retraction operation: 
\begin{equation} \label{eq:6}
    \mathfrak{R}_{\Theta^{(k)}}(\Xi)=(\Theta^{(k)}+\Xi)(I+\Xi^T \Xi)^{-\frac{1}{2}}
\end{equation}
\begin{equation}
    \mathfrak{R}_{\Theta^{(k)}}(\Xi)^T\mathfrak{R}_{\Theta^{(k)}}(\Xi)=I
\end{equation}
i.e., $\mathfrak{R}_{\Theta^{(k)}}(\Xi)$ is on $St(p,n)$ and the second step is:

\noindent
\textbf{Step 2:} Map $\Xi$ to $St(p,n)$ via Eq. \ref{eq:6}. Then the updated parameter matrix under orthogonal constraints is:

\begin{equation}
    \Theta^{(k+1)}=\mathfrak{R}_{\Theta^{(k)}}(\Theta^{(k)}-\gamma grad\ f(\Theta^{(k)}))
\end{equation}

Fig. \ref{fig:GeometricOptimization} illustrates the process of updating the parameters on the manifold. The proofs of Theorems 1 and 2 are detailed in the \red{Supplementary Material}.
\vspace*{-5pt}

\begin{figure}
    \centering
    \includegraphics[width=\linewidth]{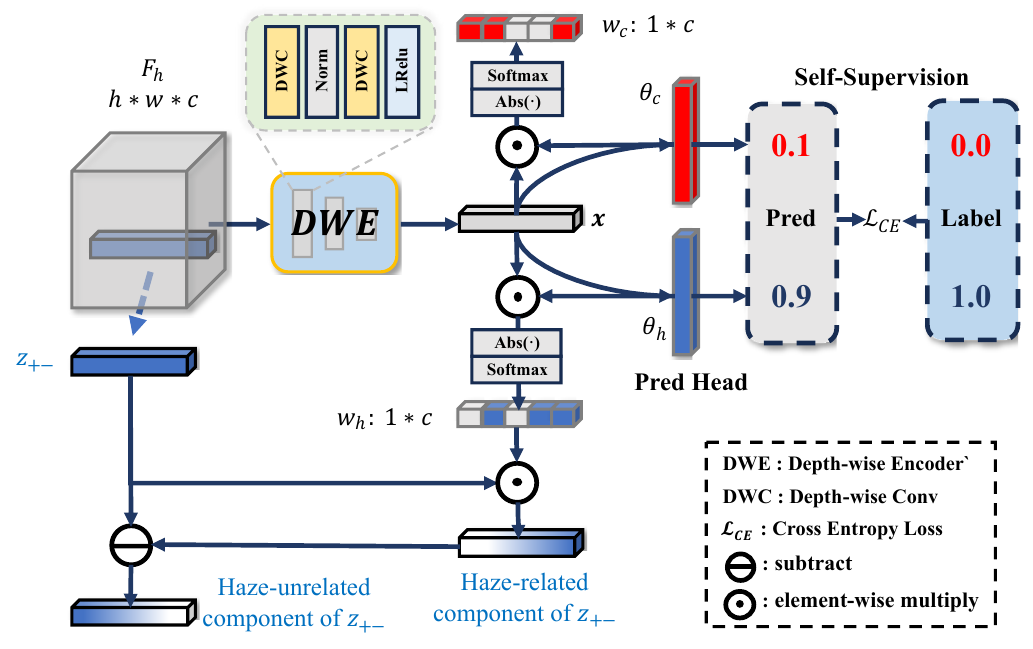}
    \caption{The structure of DWFC and the procedure for obtaining the heat-vector describing the haze relevance of features.}\label{fig:DWFC}
\end{figure}

\subsubsection{Depth-wise Feature Classifier} \label{Sec:3.3.2}
\vspace*{-5pt}
Provided the features projected into orthogonal space, it is not yet known which are haze-related and which are haze-unrelated. To solve this problem, we introduce a Depth-wise Feature Classifier (DWFC). DWFC takes the image features of $H$ or $C$ extracted by $\mathcal{G}_{enc}$ as input to predict whether the feature source is a hazy or clear image. With this self-supervised approach, one channel-wise heat-vector can be obtained for each set of input features.

Fig. \ref{fig:DWFC} illustrates the structure of DWFC. Given an orthogonal feature, it is input into the Depth-wise Encoder (DWE) and processed through 3 convolutional blocks. The processed feature is globally average pooled (GAP) to obtain a one-dimensional feature vector. The feature vector is fed to a fully connection (FC) layer to get the final classification prediction probability. Note that we use depth-wise convolution to avoid information exchange between channels to ensure that each value of the feature vector is only relevant to the corresponding channel.

Since the source of the feature is known, we use it as a label and take cross-entropy loss as a objective function to optimize the DWFC:
\begin{equation}
    \begin{split}
        \mathcal{L}_{_{CE}}=&y_hlog(\theta_h^Tx)+(1-y_h)log(1-\theta_h^Tx)\\
        & +y_clog(\theta_c^Tx)+(1-y_c)log(1-\theta_c^Tx)
    \end{split}
\end{equation}
where $y_h$ and $y_c$ denote the labels of feature source. If the source of input feature is hazy image, then $y_h=1$ and $y_c=0$. $\theta_h$ and $\theta_c$ represent the weights of the fully connection layer (pred head) of DWFC, and $x$ represents the 1-D feature vector. Thus $\theta_h^Tx$ and $\theta_c^Tx$ stand for the prediction that the source of the feature is a hazy or clear image.

Inspired by visualization methods in high-level computer vision tasks \cite{zhou2016learning,selvaraju2017grad,wang2020score}, we argue that the absolute values in the results of the element-wise multiplication between $x$ and $\theta_h$ (or $\theta_c$) reflect the magnitude of the role played by the feature of the corresponding channel in the network's decision that the source of the features is hazy (or clear). Thus the heat-vectors describing the haze (or clear) relevance can be formulated as:
\begin{equation}
    w_h=softmax(abs(\theta_h \odot x))
\end{equation}
\begin{equation}
    w_c=softmax(abs(\theta_c \odot x))
\end{equation}
where $abs(\cdot)$ denotes a function taking absolute values for all elements in the input vector, and $softmax(\cdot)$ stands for the softmax function.


For example, if the absolute value of an element in $\theta_h \odot x$ is large, it can be assumed that the feature of the corresponding channel prompt (or inhibit) the network's judgment that the feature source is a hazy image, i.e., the feature of the channel is inclined to be a hazy-related (or unrelated) feature. Specifically, for features from hazy (or clear) images, we assign them with $w_h$ (or $w_c$).

\begin{table*}[ht]\scriptsize
    \caption{Quantitative comparison of ODCR with the state-of-the-art image dehazing methods on several datasets. Best results are \textbf{bolded} and second best results are \underline{underlined}. Cells where results are not available are replaced by "-". The latency is measured on $256\times256$ images using a single RTX 4090 GPU.}\label{tab:result_synthetic}
    \begin{center}
        \vspace*{-10pt}
    \begin{tabular}{c c c c c c c c c c}
    \toprule[1pt]
    \multicolumn{2}{c}{\multirow{2}{*}{Method}} & \multicolumn{2}{c}{SOTS-indoor \cite{li2018benchmarking}} & \multicolumn{2}{c}{SOTS-outdoor \cite{li2018benchmarking}} & \multicolumn{2}{c}{NH-HAZE 2 \cite{ancuti2021ntire}} & \multicolumn{2}{c}{Overhead}\\ \cmidrule(l){3-4} \cmidrule(l){5-6} \cmidrule(l){7-8} \cmidrule(l){9-10}
    \multicolumn{2}{c}{} & PSNR (dB) & SSIM & PSNR (dB) & SSIM & PSNR (dB) & SSIM  & \#Param (M) & Latency (ms) \\
    \midrule
    \multirow{4}{*}{Paired} & DehazeNet \cite{cai2016dehazenet} & 19.82 & 0.818 & 24.75 & 0.927 & 10.62 & 0.521 & 0.009 & 0.919 \\
    & AOD-Net \cite{li2017aod} & 20.51 & 0.816 & 24.14 & 0.920 & 12.33 & 0.631 & 0.002 & 0.390 \\
    & MSCNN \cite{ren2016single} & 19.84 & 0.833 & 14.62 & 0.908 & 11.74 & 0.566 & 0.008 & 0.619 \\
    & GDN \cite{Liu_2019_ICCV} & \textbf{32.16} & \textbf{0.983} & 17.69 & 0.841 & 12.04 & 0.557 & 0.956 & 9.905 \\ \midrule
    
    \multirow{10}{*}{Unpaired} & DCP \cite{he2010single} & 13.10 & 0.699 & 19.13 & 0.815 & 14.90 & 0.668 &  - & - \\
    & CycleGAN \cite{zhu2017unpaired} & 21.34 & 0.898 & 20.55 & 0.856 & 13.95 & 0.689 & 11.38 & 10.22 \\
    & CycleDehaze \cite{engin2018cycle} & 20.11 & 0.854 & 21.31 & 0.899 & 14.12 & 0.701 & 11.38 & 10.19 \\
    & YOLY \cite{li2021you} & 15.84 & 0.819 & 14.75 & 0.857 & 13.38 & 0.595 & 32.00 & - \\
    & USID-Net \cite{li2022usid} & 21.41 & 0.894 & 23.89 & 0.919 & 15.62 & 0.740 & 3.780 & 31.01 \\
    & RefineDNet \cite{zhao2021refinednet} & 24.36 & 0.939 & 19.84 & 0.853 & 14.20 & 0.754 & 65.80 & 248.5 \\
    & $D^4$ \cite{yang2022self} & 25.42 & 0.932 & \underline{25.83} & \underline{0.956} & 14.52 & 0.709 & 10.70 & 28.08 \\
    & CUT \cite{park2020contrastive} & 24.30 & 0.911 & 23.67 & 0.904 & \underline{15.92} & \underline{0.758} & 11.38 & 10.06 \\
    \cmidrule{2-10}
    & \cellcolor{LightGray} ODCR (ours) &\cellcolor{LightGray} \underline{26.32} &\cellcolor{LightGray} \underline{0.945} &\cellcolor{LightGray} \textbf{26.16} &\cellcolor{LightGray} \textbf{0.960} &\cellcolor{LightGray} \textbf{17.56} &\cellcolor{LightGray} \textbf{0.766} &\cellcolor{LightGray} 11.38 &\cellcolor{LightGray} 10.14 \\
    \bottomrule
    \end{tabular}
    \vspace*{-15pt}
    \end{center}
    \end{table*}

\subsection{Weighted PatchNCE} 
Based on the sample partition in Sec. \ref{Sec:3.2} and the heat-vectors in Sec. \ref{Sec:3.3.2}, Weighted PatchNCE (WPNCE) for UID is proposed. WPNCE is a loss function based on mutual information between features, and we first give the definition of the weighted mutual information of two feature vectors:
\begin{equation}
    l(w, z_1, z_2)=exp(w\odot {z_2}^T\times z_1/\tau)
\end{equation}
where $w$ is the weight, $\tau$ is the temperature coefficient and $z_1$ and $z_2$ represent the two feature vectors for computing the mutual information. For WPNCE, the mutual information between the query patch and the positive components of all other key patches are desired to be maximized, which can be denoted as:
\begin{equation}
    \mathcal{P} = l(w_{h},z,z_{_{+-}})+\sum_{n = 1}^{N_{_{-+}}}l(w_{c},z,{z_{_{-+}}}^n)
\end{equation}
and minimize the mutual information with the negative components:
\begin{equation}
    \begin{split}
        \mathcal{N} =& l((\textbf{1}-w_{h}),z,z_{_{+-}})+ \sum_{n = 1}^{N_{_{--}}}l(\textbf{1},z,{z_{_{--}}}^n)\\
        & +\sum_{n = 1}^{N_{_{-+}}}l((\textbf{1}-w_{c}),z,{z_{_{-+}}}^n)
    \end{split}
\end{equation}
where $\textbf{1}$ is a vector of the same shape as $w_c$ or $w_h$ with all elements 1. Finally, we integrate them into a loss function in the form of InfoNCE.
\begin{equation}
    \mathcal{L} _{_{WPNCE}}=-log(\frac{\mathcal{P} }{\mathcal{P} +\mathcal{N} })
\end{equation}
And the full objective is as follows:

\begin{equation}
    \mathcal{L}=\mathcal{L}_{_{GAN}}+\mathcal{L}_{_{WPNCE}}+\mathcal{L}_{_{CE}}+\mathcal{L}_{idt}
\end{equation}
where $\mathcal{L}_{_{GAN}}$ and $\mathcal{L}_{idt}$ stand for the GAN loss and identity loss in CUT \cite{park2020contrastive}.

\section{Experiments}
    
\subsection{Datasets and Metrics}

We conduct experiments on several datasets to evaluate the performance of our method on UID. The datasets include RESIDE \cite{li2018benchmarking}, NH-HAZE 2 \cite{ancuti2021ntire} and Fattal's \cite{fattal2014dehazing}. The test sets cover synthetic, artificial, and real-world images.

RESIDE is a widely used image dehazing dataset containing several subsets. We use ITS (13990 pairs of indoor images) from RESIDE as the training set and SOTS (500 indoor and 500 outdoor image pairs) as the test set. NH-HAZE 2 is an artificial dataset for the NTIRE 2021 competition, which consists of 25 pairs of non-homogeneous hazy images and clear images. And Fattal's dataset is a real-world dataset that includes 41 real hazy images in various scenes. Commonly used image quality evaluation metrics: PSNR (dB) and SSIM are employed to evaluate the dehazing performance of ODCR.

\begin{figure*}
\centering
\includegraphics[width=0.9\linewidth]{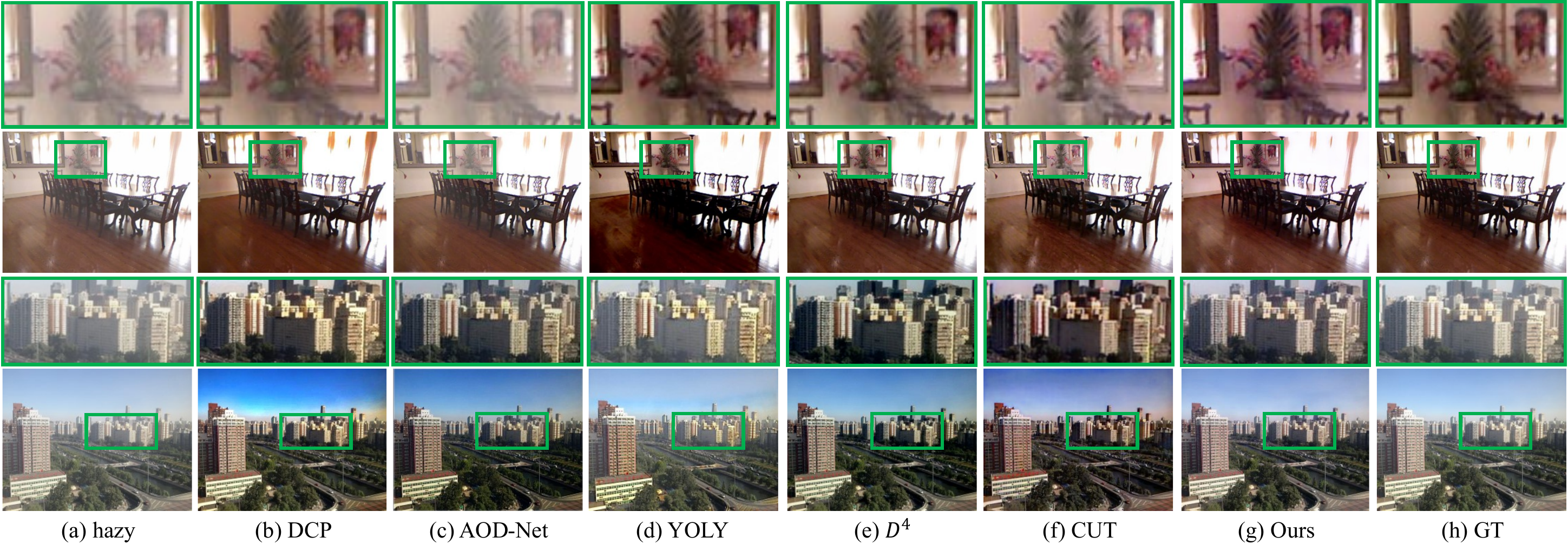}
\vspace*{-10pt}
\caption{Visual comparison of various dehazing methods on SOTS-indoor \cite{li2018benchmarking} and SOTS-outdoor \cite{li2018benchmarking}. Areas where our method works better are boxed out and zoomed in, or you can zoom in by yourself to get a better view.}\label{fig:Visual_compare}
\vspace*{-10pt}
\end{figure*}

\begin{figure*}
\centering
\includegraphics[width=0.9\linewidth]{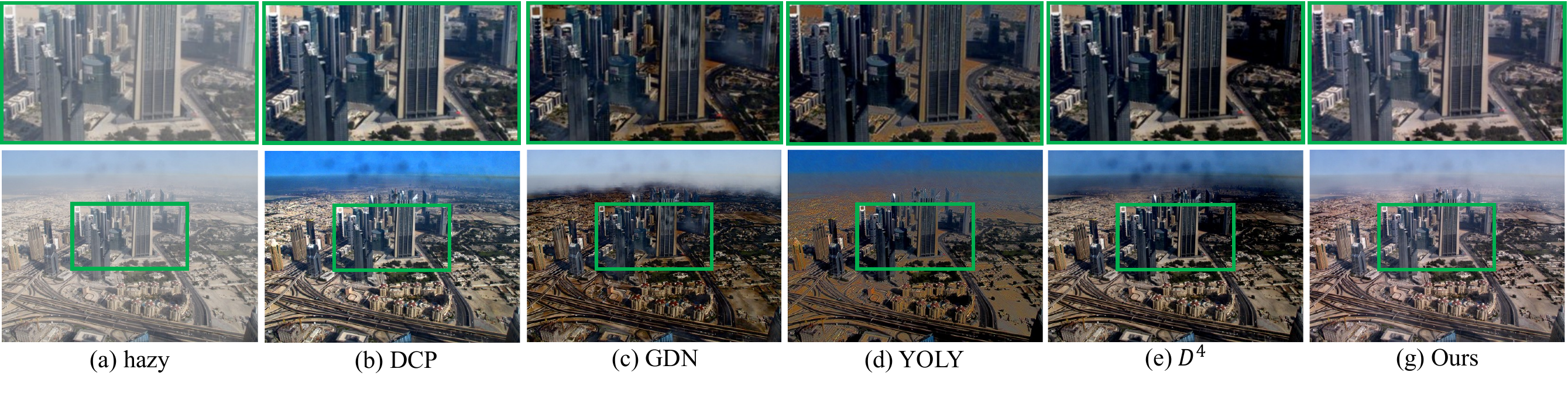}
\vspace*{-10pt}
\caption{Visual comparison of various dehazing methods on Fattal's dataset \cite{fattal2014dehazing}. Areas where our method works better are boxed out and zoomed in, or you can zoom in by yourself to get a better view.}\label{fig:Real_compare}
\vspace*{-15pt}
\end{figure*}

\subsection{Performance Evaluation}


We compare ODCR with several SOTA image dehazing methods. Some of them are trained using paired data, including DehazeNet \cite{cai2016dehazenet}, AOD-Net \cite{li2017aod}, MSCNN \cite{ren2016single} and GDN \cite{Liu_2019_ICCV}. Others do not require paired data, including DCP \cite{he2010single}, CycleGAN \cite{zhu2017unpaired}, CycleDehaze \cite{engin2018cycle}, YOLY \cite{li2021you}, USID-Net \cite{li2022usid}, RefineDNet \cite{zhao2021refinednet}, $D^4$ \cite{yang2022self} and CUT \cite{park2020contrastive}. Notably, we follow the evaluation strategy of $D^4$ \cite{yang2022self}. We train our model only on ITS and test on all the test sets and all other methods also follow this strategy for fairness.

\noindent
\textbf{Quantitative Evaluation.} The quantitative comparison of different methods is recorded in Table \ref{tab:result_synthetic}. Our ODCR obtains the second-best results on the SOTS-indoor dataset, and the best results are obtained by GDNet, which is a supervised method. However, testing the same model on the SOTS-outdoor and NH-HAZE 2 datasets, which are not trained accordingly, ODCR outperforms all paired and unpaired data-based methods, including GDNet. The results above demonstrate that ODCR is able to learn the patterns of haze-free images with excellent generalization by decoupling the image features into haze-related and unrelated components and performing contrastive learning separately.

\noindent
\textbf{Qualitative Evaluation.} The qualitative comparison of synthetic and artificial datasets between various methods is displayed in Fig. \ref{fig:Visual_compare}. The supervised method AOD-Net \cite{li2017aod} fails to dehaze in local area. DCP \cite{he2010single}, YOLY \cite{li2021you} and $D^4$ \cite{yang2022self} suffer from color distortion. Compared with CUT \cite{park2020contrastive}, which only performs contrastive learning on the hazy patches, ODCR shows improved overall dehazing performance while maintaining haze-unrelated information.

In addition, the visual comparison of the different methods on the real-world dataset Fattal's is shown in Fig. \ref{fig:Real_compare}. It obviously illustrates that ODCR better removes haze from the boxed region compared to other methods and has minimal image quality degradation including artifacts, loss of structural details, and color distortion, verifying the dehazing effectiveness of ODCR on real-world images.

\subsection{Ablation Studies and Discussion}

\textbf{Effectiveness of Orthogonal Decoupling.}
We propose an O-MLP with orthogonal constraint to minimize feature embedding relevance, effectively decoupling haze-related and unrelated features. Our quantitative analysis contrasts three distinct cases: the absence of Orthogonal Decoupling (OD), the imposition of an orthogonal loss function, and optimization conducted on the Stiefel manifold. The comparative results, as detailed in Table \ref{tab:OD}, affirm that the Stiefel manifold optimization outperforms the other approaches in dehazing performance.

\begin{figure}[bp]
        \centering
        \includegraphics[width=\linewidth]{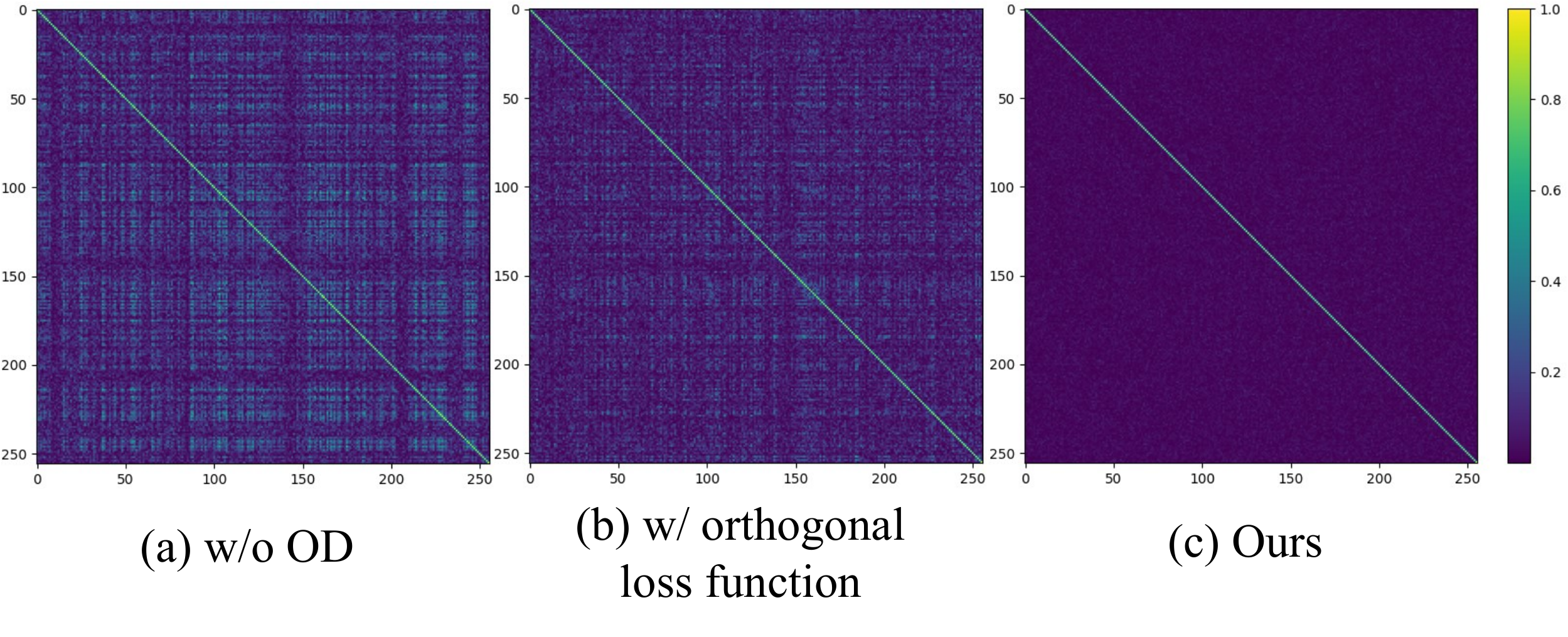}
        \vspace*{-20pt}
        \caption{Visualization of the similarity matrix with different settings. }\label{fig:cos_corr}
    \end{figure}

To further substantiate the O-MLP's capacity to attenuate feature relevance, we examine the cosine similarity matrices of the feature embeddings. Fig. \ref{fig:cos_corr} delineates the inter-channel cosine similarity matrices derived from the feature projections in each case. It indicates a substantial redundancy in features without OD. While the orthogonal loss function offers a reduction in feature relevance, it does not achieve optimal separation. In stark contrast, O-MLP, when optimized on the Stiefel manifold, demonstrates a marked minimization in feature relevance, which we attribute to its strict orthogonality. This geometric optimization is pivotal in achieving the desired feature decoupling necessary for effective dehazing.

\begin{table}[bp]
    \caption{Ablation study on the effectiveness of O-MLP.}
    \begin{center}
        \vspace*{-10pt}
    \begin{tabular}{c c c}
    \toprule[1pt]
    \multirow{2}{*}{Setting} & \multicolumn{2}{c}{Metric}\\ \cmidrule(l){2-3}
    & PSNR (dB) & SSIM \\
    \midrule
    w/o OD & 22.30 & 0.887 \\
    \midrule
    w/ orthogonal loss & 24.96 & 0.928 \\
    \midrule
    \rowcolor{LightGray} w/ O-MLP (ours) & \textbf{26.32} \myfontsize (\textcolor{green}{+1.36}) & \textbf{0.945} \myfontsize (\textcolor{green}{+0.017}) \\
    \bottomrule
    \end{tabular}
    \vspace*{-10pt}
    \end{center}
    \label{tab:OD}
    \end{table}

\begin{table}[bp]
    \caption{Ablation study on the feature assignment method. Note that Ratio in the table represents assigning orthogonal features in a related:unrelated ratio. For example, when ratio is 1:7, the first 32 channels of a 256-channel orthogonal feature are assigned as related features, and the last 224 channels are unrelated features.}
    \begin{center}
        \vspace*{-10pt}
    \begin{tabular}{c c c c}
    \toprule[1pt]
    \multirow{2}{*}{Method} & \multirow{2}{*}{Ratio} & \multicolumn{2}{c}{Metric}\\ \cmidrule(l){3-4}
    & & PSNR (dB) & SSIM\\
    \midrule
    
    \multirow{5}{*}{Fixed} & 1:7 & 13.10 & 0.599 \\
    & 1:3 & 13.93 & 0.574 \\
    & 1:1 & 15.33 & 0.621 \\
    & 3:1 & 14.30 & 0.660 \\
    & 7:1 & 18.28 & 0.726 \\
    \midrule
    \rowcolor{LightGray} DWFC (ours) & N/A & \textbf{26.32} \myfontsize (\textcolor{green}{+8.04}) & \textbf{0.945} \myfontsize (\textcolor{green}{+0.219})\\
    \bottomrule
    \end{tabular}
    \vspace*{-10pt}
    \end{center}
    \label{tab:DWFC}
    \end{table}

\begin{figure}
    \centering
    \includegraphics[width=\linewidth]{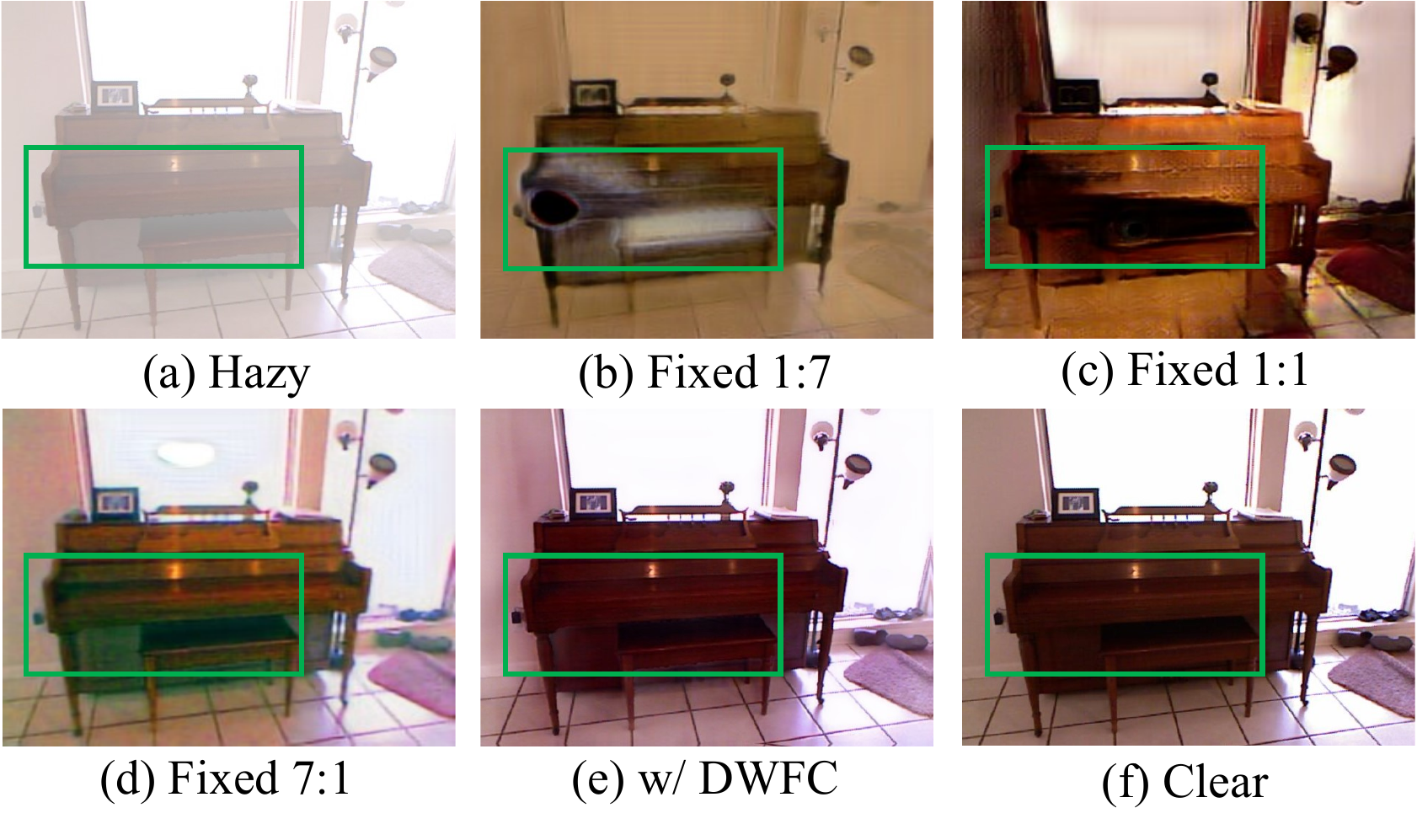}
    \vspace*{-15pt}
    \caption{Qualitative comparison on the assignment of related/unrelated features. (a) is the hazy input. (b)-(d) are the dehazing results of the model with different ratio of assignment. (e) is the result of the default network. And (f) is the corresponding clear image.}\label{fig:ablation_DWFC}
    \vspace*{-15pt}
\end{figure}

\noindent
\textbf{Effectiveness of DWFC.} 
We evaluate the effectiveness of the proposed DWFC. DWFC assigns haze-related and unrelated features through a self-supervised mechanism leveraging the heat-vector output. Another feasible way is to statically determine a fixed ratio of orthogonal features as haze-related or unrelated. As Table \ref{tab:DWFC} demonstrates, irrespective of the predetermined ratio of related to unrelated features, the dehazing capability is significantly compromised. Fig. \ref{fig:ablation_DWFC} also presents a qualitative comparison of dehazing outcomes under different configurations. Fixed feature ratios lead to a notable decline in image quality. The degradations above stem from the absence of the DWFC's self-supervision, causing features unrelated to haze that should be mapped to the hazy domain to be mistakenly aligned with the clear domain during contrastive learning. Consequently, this misalignment distorts information channels, such as texture and semantics, critical for accurate dehazing. In addition, the gradual improvement of the dehazing performance as the percentage of haze-unrelated feature increases in Table \ref{tab:DWFC} and Fig. \ref{fig:ablation_DWFC} also proves the above view.

\begin{figure}
    \centering
    \includegraphics[width=0.8\linewidth]{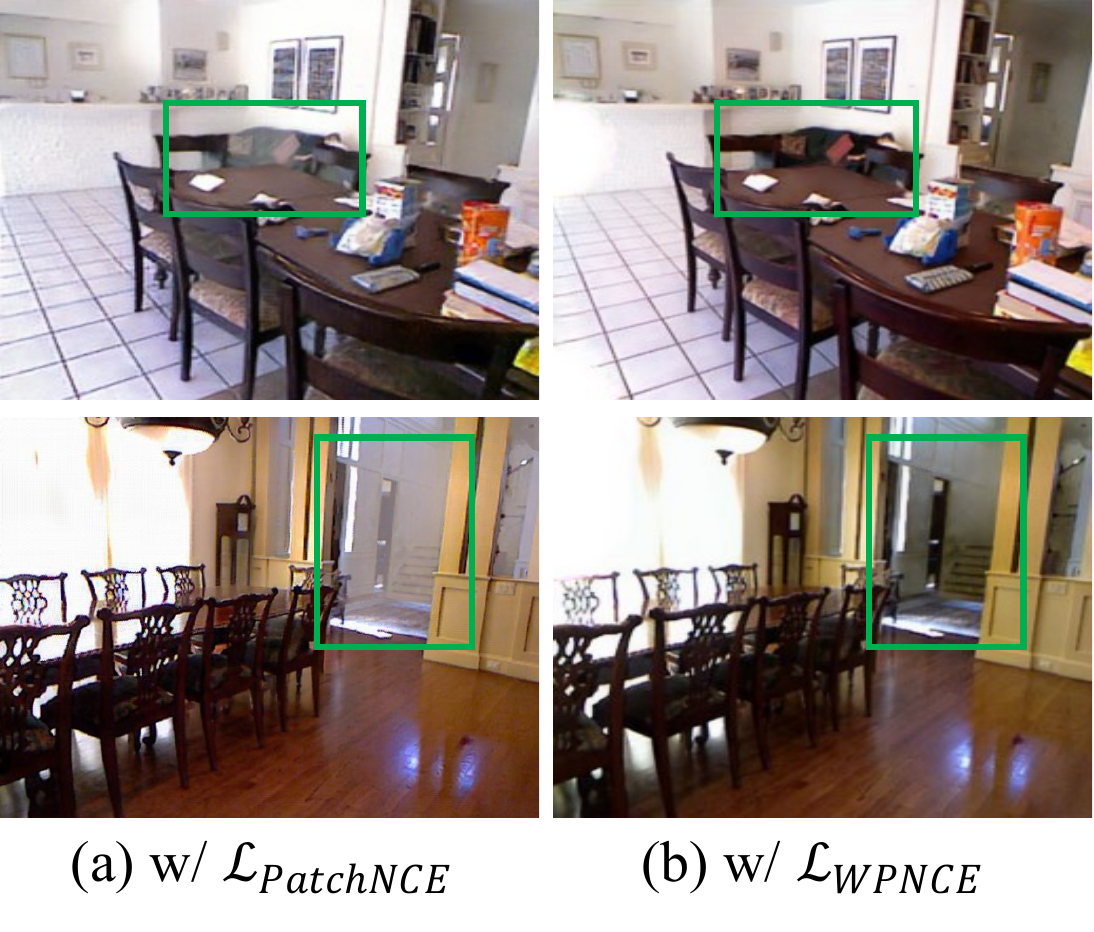}
    \caption{Qualitative comparison of PatchNCE and WPNCE.}\label{fig:ablation_WPNCE}
\end{figure}

\noindent
\textbf{Effectiveness of WPNCE.} Experiments are conducted on the loss functions to verify the effectiveness of our proposed WPNCE. Compared to PatchNCE, WPNCE exploits the haze-related component in clear images to drive the generated images close to being clear. The quantitative and qualitative comparison between the results of using PatchNCE and WPNCE is recorded in Table \ref{tab:WPNCE} and Fig. \ref{fig:ablation_WPNCE}, respectively. Better dehazing results are obtained using WPNCE compared to using PatchNCE, due to the fact that WPNCE treats haze-related and unrelated features differently and avoids the contradiction of dehazing and maximizing the mutual information between the generated image features and the haze patch features.

\begin{table}[ht]
\caption{Ablation study on the effectiveness of $\mathcal{L}_{_{WPNCE}}$.}
\begin{center}
    \vspace*{-10pt}
\begin{tabular}{c c c}
\toprule[1pt]
\multirow{2}{*}{Setting} & \multicolumn{2}{c}{Metric}\\ \cmidrule(l){2-3}
& PSNR (dB) & SSIM \\
\midrule
$\mathcal{L}_{_{PatchNCE}}$ \cite{park2020contrastive} & 24.30 & 0.911 \\
\midrule
\rowcolor{LightGray} $\mathcal{L}_{_{WPNCE}}$ (ours) & \textbf{26.32} \myfontsize (\textcolor{green}{+2.02}) & \textbf{0.945} \myfontsize (\textcolor{green}{+0.034}) \\
\bottomrule
\end{tabular}
\vspace*{-20pt}
\end{center}
\label{tab:WPNCE}
\end{table}

\section{Conclusion}

We propose Orthogonal Decoupling Contrastive Regularization for UID by decoupling image feature into haze-related/-unrelated components. In particular, we repartition the patch samples in terms of haze-related/-unrelated. Afterward, an orthogonal MLP geometrically optimized on the Stiefel manifold is introduced to reduce the relevance between the features by projecting the features of each sample into the orthogonal space. Furthermore, a Depth-wise Feature Classifier for assigning the projected features of each channel as haze-related/-unrelated is proposed. Finally, a novel Weighted PatchNCE is designed to maximize the mutual information between the corresponding components of query and positive samples in different feature spaces. Experiments conducted on synthetic and real-world datasets validate our proposal and analysis. Performance improvements are observed when comparing other SOTA methods for UID.

\noindent \\
\textbf{Acknowledgement.} 

\noindent This work was supported by the National Natural Science Foundation
of China (NSFC) under Grant 62173143.
{
    \small
    \bibliographystyle{ieeenat_fullname}
    \bibliography{main}
}


\end{document}


\appendix
\maketitle

\section{Implementation Details} 
We follow the setting of CUT \cite{park2020contrastive}, except that the PatchNCE loss is repalced by our Weighted PatchNCE (WPNCE). In detail, we use a 9-block Resnet-based generator \cite{johnson2016perceptual} with PatchGAN \cite{isola2017image}. We define our encoder as the first half of the generator, and accordingly extract our multilayer features from five evenly distributed points of the encoder. For the O-MLPs, we set their numbers of layers to 2 with 256 units in each layer and $\tau$ to 0.07.

We train our full model for 400 epochs, keeping the learning rate at $2e^{-4}$ for the first 200 epochs, and linearly decay the learning rate in the last 200 epochs. Each subnet of the network is optimized by the Adam optimizer \cite{kingma2014adam} ($\beta_1=0.5, \beta_2=0.999$). The samples are randomly cropped into patches of size $256\times 256$ during training.

\section{Proofs of Theorems}
\textbf{Theorem 1.} Given $\nabla f(\cdot )$ in Euclidean space and $grad\ f(\cdot)$ defined as follows:
\begin{equation}\label{eq:1}
    grad\ f(\Theta)=\nabla f(\Theta)-\frac{1}{2}\Theta\Theta^T\nabla f(\Theta)-\frac{1}{2}\Theta{\nabla f(\Theta)}^T\Theta
\end{equation}
$grad\ f(\Theta)$ is the orthogonal projection of $\nabla f(\Theta)$ onto the tangent space of the Stiefel manifold.

\noindent
\textbf{Proof 1.} If $grad\ f(\cdot)$ is the orthogonal projection of $\nabla f(\cdot)$ onto the tangent space of the Stiefel manifold, according to the triangle law of vector addition, we have:
\begin{equation}\label{eq:2}
    \langle (grad\ f(\cdot)-\nabla f(\cdot)),grad\ f(\cdot)\rangle =0
\end{equation}
where $\langle\cdot,\cdot\rangle$ is the inner product. In particular, for $Z_1$, $Z_2$ on the Stiefel manifold:
\begin{equation}\label{eq:3}
    \langle Z_1,Z_2\rangle=tr(Z_1^TZ_2^T)
\end{equation}

By substituting Eq. \ref{eq:1} and \ref{eq:3} into the left-hand side of Eq. \ref{eq:2}, we obtain:
{
\footnotesize
\begin{equation}\label{eq:4}
    \begin{split}
        &\langle (grad\ f(\Theta)-\nabla f(\Theta)),grad\ f(\Theta)\rangle\\
        =&tr(\frac{1}{2}\nabla f(\Theta)^T\Theta\Theta^T\nabla f(\Theta)-\frac{1}{4}\nabla f(\Theta)^T\Theta\Theta^T\Theta\Theta^T\nabla f(\Theta)\\
        &-\frac{1}{4}\nabla f(\Theta)^T\Theta\Theta^T\Theta \nabla f(\Theta)^T\Theta+\frac{1}{2}\Theta^T\nabla f(\Theta)\Theta^T\nabla f(\Theta)\\
        &-\frac{1}{4}\Theta^T\nabla f(\Theta)\Theta^T\Theta\Theta^T\nabla f(\Theta)-\frac{1}{4}\Theta^T\nabla f(\Theta)\Theta^T\Theta\nabla f(\Theta)^T\Theta)
    \end{split}
\end{equation}}
Due to $\Theta\in St(p,n)$, $\Theta$ satisfies $\Theta^T\Theta=I$. Then Eq. \ref{eq:4} can be simplify as:
{\footnotesize
\begin{equation}\label{eq:6}
    \begin{split}
        &\langle (grad\ f(\Theta)-\nabla f(\Theta)),grad\ f(\Theta)\rangle\\
        =&\frac{1}{4}tr(\nabla f(\Theta)^T\Theta\Theta^T\nabla f(\Theta))-\frac{1}{4}tr(\nabla f(\Theta)^T\Theta\nabla f(\Theta)^T\Theta)\\
        &+\frac{1}{4}tr(\Theta^T\nabla f(\Theta)\Theta^T\nabla f(\Theta))-\frac{1}{4}tr(\Theta^T\nabla f(\Theta)\nabla f(\Theta)^T\Theta)
    \end{split}
\end{equation}}
Due to $tr(AB)=tr(BA)$ and $tr(A)=tr(A^T)$, we have
\begin{equation}
    tr(\nabla f(\Theta)^T\Theta\Theta^T\nabla f(\Theta))=tr(\Theta^T\nabla f(\Theta)\nabla f(\Theta)^T\Theta)
\end{equation}
and
\begin{equation}
    tr(\Theta^T\nabla f(\Theta)\Theta^T\nabla f(\Theta))=tr(\nabla f(\Theta)^T\Theta\nabla f(\Theta)^T\Theta)
\end{equation}
So Eq. \ref{eq:6} equals to 0, demonstrating $grad\ f(\Theta)$ is the orthogonal projection of $\nabla f(\Theta)$. Furthermore, it can also prove that $\Theta^Tgrad\ f(\Theta) + grad\ f(\Theta)^Tf(\Theta)=0$, which means that $grad\ f(\Theta)$ is on the tangent space.

\noindent
\textbf{Theorem 2.} Given a point $\Xi=\Theta-\gamma grad\ f(\Theta)$ on the tangent space $T_{\Theta} St(p,n)$ of a point on a Stiefel manifold, let retraction operation be $\mathfrak{R}_{\Theta}(\Xi)=(\Theta+\Xi)(I+\Xi^T \Xi)^{-\frac{1}{2}}$, the point $\Xi$ after retraction operation satisfies orthogonal constraint:
\begin{equation}
    \mathfrak{R}_{\Theta}(\Xi)^T\mathfrak{R}_{\Theta}(\Xi)=I
\end{equation}
i.e., $\mathfrak{R}_{\Theta}(\Xi)$ is on $St(p,n)$.

\noindent
\textbf{Proof 2.}
The parameter after retraction $\mathfrak{R}_{\Theta}$ satisfies:
{\scriptsize
\begin{equation}
    \mathfrak{R}_{\Theta}(\Xi)^T\mathfrak{R}_{\Theta}(\Xi)=(I+\Xi^T\Xi)^{-\frac{1}{2}}((\Theta+\Xi)^T(\Theta+\Xi))(I+\Xi^T\Xi)^{-\frac{1}{2}}
\end{equation}
}
For $\Xi$ on the tangent space $T_{\Theta} St(p,n)$, $\Xi$ satisfies:
\begin{equation}
    \Theta^T\Xi+\Xi^T\Theta=0
\end{equation}
So 
\begin{equation}
    \mathfrak{R}_{\Theta}(\Xi)^T\mathfrak{R}_{\Theta}(\Xi)=(I+\Xi^T\Xi)^{-\frac{1}{2}}(I+\Xi^T\Xi)(I+\Xi^T\Xi)^{-\frac{1}{2}}
\end{equation}
where $(I+\Xi^T\Xi)\in\mathbb{R} ^{p\times p}$. Eigenvalue decomposition of $(I+\Xi^T\Xi)$ can get:
\begin{equation}
    (I+\Xi^T\Xi)=P\varLambda P^{-1}
\end{equation}
where $\varLambda $ stands for a diagonal matrix, and $P^TP=I$. Subsequently, we have:
\begin{equation}
\begin{split}
        \mathfrak{R}_{\Theta}(\Xi)^T\mathfrak{R}_{\Theta}(\Xi)&=P\varLambda ^{-\frac{1}{2}}P^TP\varLambda P^TP\varLambda ^{-\frac{1}{2}}P^T\\
        &=P\varLambda ^{-\frac{1}{2}}\varLambda \varLambda ^{-\frac{1}{2}}P^T\\
        &=PP^T\\
        &=I
\end{split}
\end{equation}
Thus $\mathfrak{R}_{\Theta}(\Xi)$ satisfies the orthogonal constraint.

        
{
    \small
    \bibliographystyle{ieeenat_fullname}
    \bibliography{supple}
}
